\documentclass[preprint,12pt]{elsarticle}

\usepackage{amssymb}
\usepackage{algorithm}
\usepackage{algpseudocode}
\usepackage{graphicx}
\usepackage{xspace}
\usepackage[dvipsnames]{xcolor}

\usepackage{hyperref}       % hyperlinks
\usepackage{url}            % simple URL typesetting
\usepackage{booktabs}       % professional-quality tables
\usepackage{amsfonts}       % blackboard math symbols
\usepackage{nicefrac}       % compact symbols for 1/2, etc.
\usepackage{microtype}      % microtypography
\usepackage{lipsum}
\usepackage{fancyhdr}       % header
\usepackage{graphicx}       % graphics
\graphicspath{{media/}}     % organize your images and other figures under media/ folder
\usepackage{amsmath}
\usepackage{amssymb}
\usepackage{breqn}
\usepackage{paralist}
\usepackage{xcolor}
\usepackage{subfig}
\usepackage{comment}
\usepackage{appendix}
\usepackage{multirow}
\usepackage{bbm}
\usepackage[left]{lineno}

\newcommand{\frcnn}{Faster R-CNN\xspace}
\newcommand{\fcos}{FCOS\xspace}

\newcommand{\detr}{DETR\xspace}
\newcommand{\bboxcut}{BBoxCut\xspace}

\journal{Computers and Electronics in Agriculture}

\begin{document}
%\linenumbers

\begin{frontmatter}



\title{BBoxCut: A Targeted Data Augmentation Technique for Enhancing Wheat Head Detection Under Occlusions}


\author[amrita]{Yasashwini Sai Gowri P}
\author[hydronium, liat]{Karthik Seemakurthy}
\author[uol]{Andrews Agyemang Opoku}
\author[amrita]{Sita Devi Bharatula}

\affiliation[amrita]{organization={Amrita School of Engineering, Amrita Vishwa Vidyapeetham},%Department and Organization
            addressline={Department of Electronics and Communications Engineering}, 
            city={Chennai},
            country={India}}
\affiliation[hydronium]{organization={Hydronium Energies Limited},%Department and Organization
             addressline={London},
            city={London},
            country={UK}}
\affiliation[uol]{organization={School of Computer Science},%Department and Organization
            addressline={University of Lincoln}, 
            city={Lincoln},
            country={UK}}
\affiliation[liat]{organization={Lincoln Institute of Agri-Food Technology},%Department and Organization
            addressline={University of Lincoln}, 
            city={Lincoln},
            country={UK}}

\begin{abstract}
Wheat plays a critical role in global food security, making it one of the most extensively studied crops. Accurate identification and measurement of key characteristics of wheat heads are essential for breeders to select varieties for cross-breeding, with the goal of developing nutrient-dense, resilient, and sustainable cultivars. Traditionally, these measurements are performed manually, which is both time-consuming and inefficient. Advances in digital technologies have paved the way for automating this process. However, field conditions pose significant challenges, such as occlusions of leaves, overlapping wheat heads, varying lighting conditions, and motion blur. In this paper, we propose a novel data augmentation technique, \bboxcut, which uses random localized masking to simulate occlusions caused by leaves and neighboring wheat heads. We evaluated our approach using three state-of-the-art object detectors and observed mean average precision (mAP) gains of 2.76, 3.26, and 1.9 for \frcnn, \fcos, and \detr, respectively. Our augmentation technique led to significant improvements both qualitatively and quantitatively. In particular, the improvements were particularly evident in scenarios involving occluded wheat heads, demonstrating the robustness of our method in challenging field conditions. 
\end{abstract}

\begin{graphicalabstract}
\includegraphics[width=400px,height=200px]{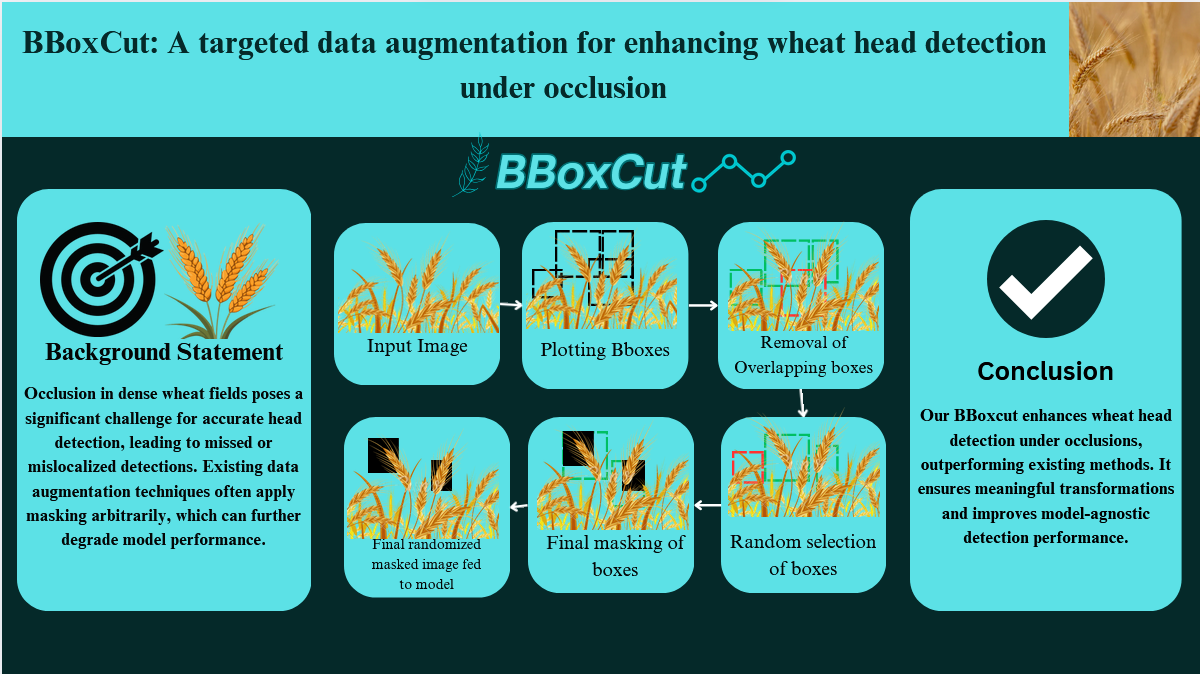}
\end{graphicalabstract}

\begin{highlights}
\item Proposed a random localised masking based data augmentation technique for improved occluded wheat detection.
\item Improved detection performance for two scenarios: a) wheat heads that overlap each other and b) wheat heads that are occluded by leaves.
\item The effectiveness of the proposed approach was evaluated across three different object detection architectures. 
\end{highlights}

\begin{keyword}
Wheat Detection \sep Occlusion \sep Data Augmentation \sep Masking


\end{keyword}

\end{frontmatter}


\section{Introduction}
Wheat is a staple crop for more than three billion people around the world and serves as a critical source of essential nutrients. To meet the increasing demands of a growing global population, agricultural production must double by 2050 \citep{Foley:2011}. Given its status as the most widely grown crop, accurate monitoring of wheat yield is crucial to ensure global food security \citep{Chakraborty:2011}. Estimating wheat yield and making informed decisions about which wheat varieties to cross-breed for more nutritious, resilient, and high-yield cultivars are dependent on accurate detection of wheat head. Traditionally, yield prediction has relied on expert forecasts and manual counting of wheat heads, which are both time consuming and prone to inaccuracies~\cite{madec:2019}.

\begin{figure}[ht]
	\centering
           \begin{tabular}{cc}
            \includegraphics[width=0.45\linewidth]{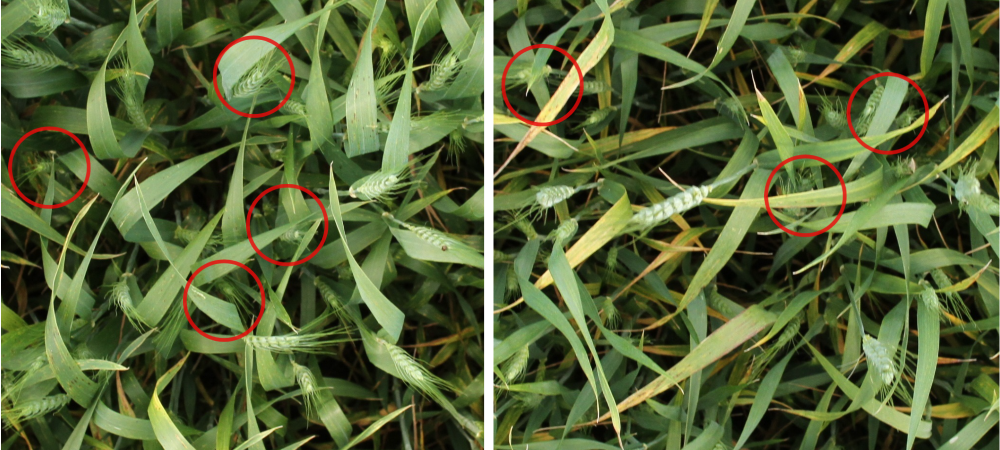} &
            \includegraphics[width=0.45\linewidth]{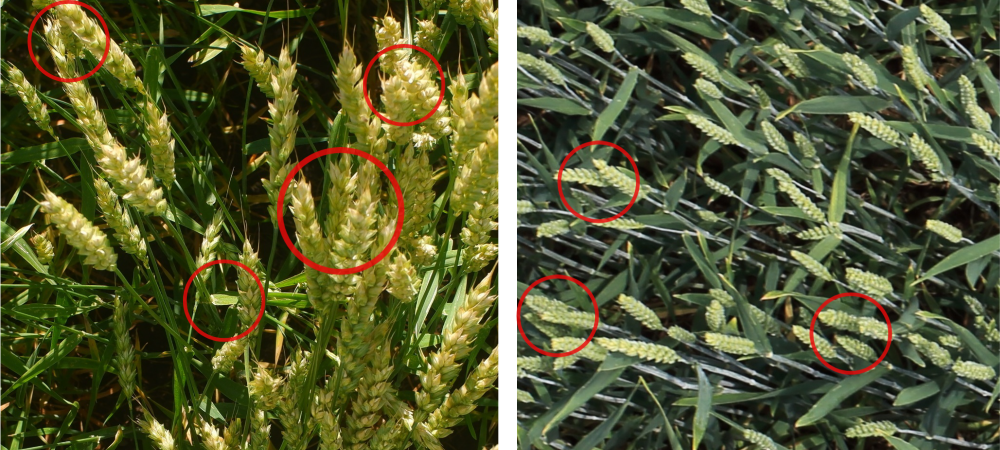}\\
            (a) & (b) \\
           \end{tabular}
	\caption{Samples from GWHD 2021 dataset. (a) Wheat heads partially occluded by leaves, (b) Wheat heads occluding each other. Occlusions are indicated by the red circles.}
	\label{fig:OcclusionDemo}
\end{figure}

\begin{figure}
    \centering
    \includegraphics[width=1.0\linewidth]{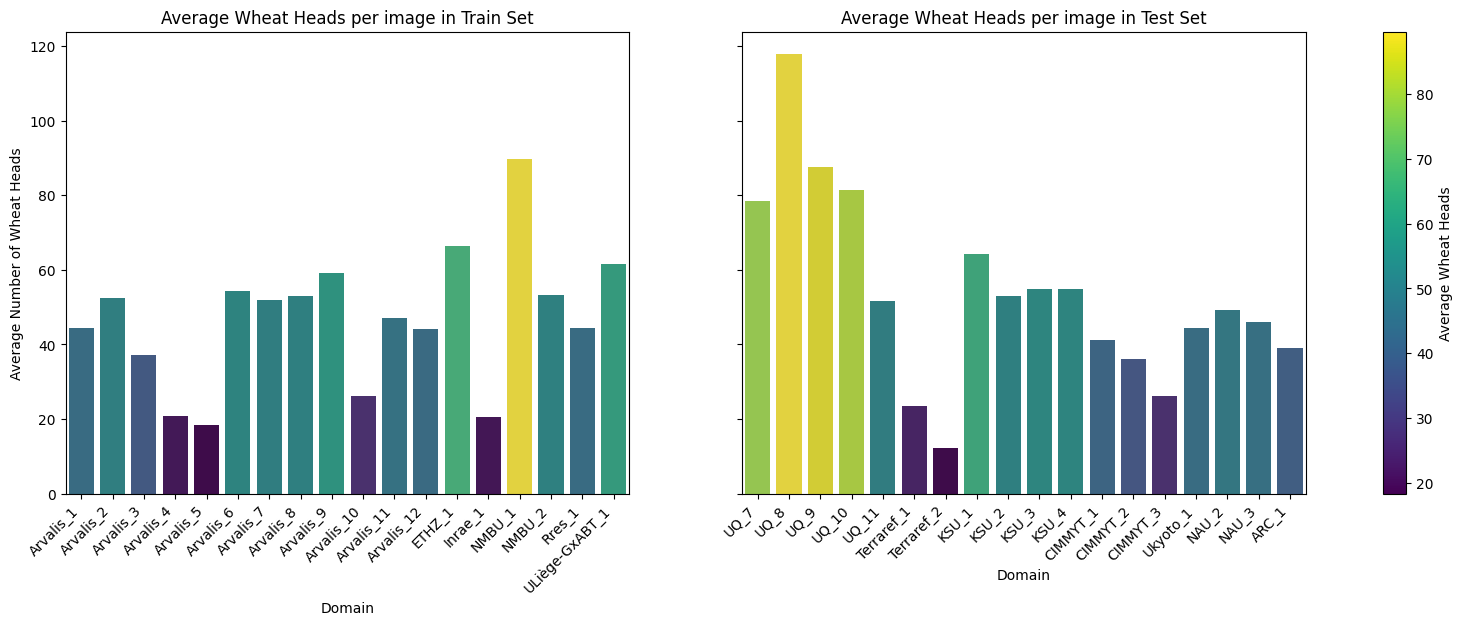}
    \caption{Average wheat head count per image in GWHD 2021 dataset. }
    \label{fig:avg_wheat_count}
\end{figure}

The Global Wheat Head Detection (GWHD) 2021 dataset~\cite{David:2021} was introduced to facilitate the development of automated systems capable of measuring the visual traits of wheat heads. Despite significant advances in object detection models~\cite{zou2023object}, detecting wheat heads in unconstrained real-world field conditions presents considerable challenges, as illustrated in Fig.~\ref{fig:OcclusionDemo}. Fig.~\ref{fig:OcclusionDemo} (a) shows occlusions caused by leaves, while Fig.~\ref{fig:OcclusionDemo} (b) highlights wheat heads that overlap one another. In both cases, state-of-the-art object detectors often struggle to accurately locate wheat heads. This performance degradation is likely due to the lack of diverse training examples that effectively represent occlusion scenarios. Furthermore, Fig.\ref{fig:avg_wheat_count} reveals that the test domains exhibit higher wheat head densities compared to the training set, further increasing the likelihood of occlusions caused by leaves and neighboring wheat heads, as shown in Fig.\ref{fig:OcclusionDemo}. Such occlusions are expected to significantly affect the average performance of detectors across unseen test domains.

We propose \bboxcut, a data augmentation technique that applies random localized masking to simulate realistic occlusions in wheat fields. Unlike Cutout \citep{DeVries:2017}, which randomly masks square regions throughout the image, \bboxcut specifically masks portions of the bounding boxes, mimicking partial occlusions by leaves or neighboring wheat heads. By incorporating \bboxcut into the data augmentation pipeline for the training of object detection models (\frcnn, \fcos, and \detr) we observed significant improvements in mean Average Precision (mAP), particularly under occlusion scenarios. This enhancement is crucial to accurately predict wheat yields, directly impacting farmer profitability. The following is a list of our contributions proposed in this paper.
\begin{itemize}
    \item We proposed a new data augmentation technique that improves wheat head localization in the presence of occlusions.
    \item We propose using a histogram-based dominant color estimation for mask generation. 
    \item We have shown the validity of our approach on three different types of detectors such as \frcnn, \fcos, and \detr.
    \item Our comparison with state-of-the-art masking-based data augmentation approaches demonstrates improved detection capability using the proposed technique.   
\end{itemize}

\section{Related work}
\label{sec:related_works}

In this section, we provide an overview of related works, taking a broad look at deep learning-based object detection, recent efforts to improve wheat detection accuracy, and studies that explicitly model occlusion in the context of object detection. We also highlight how our approach differs from existing techniques.

\subsection{Deep Learning for object detection}
\label{sec:DL_OD}
Deep neural networks have been adopted for various computer vision tasks, such as object detection, due to their exceptional performance in feature extraction \citep{Alex:2012}. Most object detection networks can be classified into two categories: single-stage~\citep{Redmon:2016, Redmon:2018, Alexey:yolo:2020, Jocher:2020, Liu:2016, tian2020fcos} and two-stage~\citep{Girshick:2015, He:2015}. Although most existing approaches use convolutional neural network (CNN) architectures, transformer-based architectures~\citep{carion2020end, gupta2022ow} have been gaining importance. Despite significant improvements in speed and accuracy, deep learning-based object detection networks generally do not explicitly model occlusion \citep{Ayvaci:2012}, which is common in many real-world scenarios~\citep{David:2021, blok2016machine}.

\subsection{Wheat Head Detection}
The importance of wheat for global food security has driven researchers to apply various detection techniques to detect wheat heads. Earlier work used image processing and machine learning techniques for wheat head detection. \cite{Narkhede:2015} used color-based features to count wheat heads. However, wheat ears, leaves, and stalks in the wheat field share similar color and texture characteristics. In addition, as the wheat plant matures, the color of different parts of the plant changes. As a result, using color cues to reliably identify wheat ears proved challenging. \cite{Zhu:2016} successfully detected and counted wheat ears using the support vector machine approach. \cite{Xu:2020} utilized the k-means approach to partition the wheat head and obtain identification. To segment and count wheat heads, the twin support vector machine segmentation approach was used in \cite{Zhou:2018}. \cite{Gallego:2018} segmented wheat heads from backgrounds using Fourier filtering and Fourier transform.

Due to the limitations of traditional object detection methods, researchers in recent years have adopted deep learning-based wheat head localization techniques. \cite{Hasan:2018} used R-CNN to detect and analyze wheat heads. Similarly, \cite{Abhi:2021} applied ablation in color, affine variation, and random cutout on the Global Wheat Head Detection (GWHD) data set and trained the model on a pre-trained YOLOv5 model. Although the performance of the model was not overly impressive, the random cutout showed an increase in performance compared to the ablation of color and the affine variation. \cite{Aksh:2021} successfully trained a ResNet-50 model to detect and classify wheat head diseases, but no mechanisms for handling occlusion were mentioned. \cite{Seemakurthy:2022} proposed a domain-independent feature representation approach that performed well when tested with wheat head data from several locations, but the problem of missed detection due to occlusion persisted.

\subsection{Occlusion Handling in Object Detection}
The presence of occlusions that hinder object visibility is one of the key factors contributing to the decrease in object detection network performance. Recently, numerous works have summarized and highlighted the need to modify occlusions to advance the state-of-the-art object detection performance~\cite{occ_survey_1, occ_survey_2, occ_survey_3, occ_survey_4, occ_survey_5}. The latest work \cite{occ_survey_5} summarized the occlusion handling approaches into three different categories: Generative techniques, deep learning strategies, and alternative approaches like graphical models and data augmentation techniques. The key advantage of data augmentation approaches over other occlusion modelling techniques is that they do not introduce additional trainable parameters, making them well-suited for resource-constrained scenarios.The proposed approach falls under the data augmentation category. In the next subsection, we describe related works that also belong to this category.

\subsubsection{Data Augmentation}
\label{sec:data_augmentation}
The quantity and diversity of training data are critical factors in determining a model's ability to generalize effectively. Data augmentation is a key strategy to generate additional useful data from existing datasets, enabling models to achieve optimal performance. For instance, \cite{Yang:2022} conducted experiments with and without data augmentation to evaluate object detection algorithms, demonstrating that models trained with augmented data consistently outperformed those trained without it. Existing data augmentation approaches can be broadly categorized into three types: spatial transformation \citep{Alex:2012}, color distortion \citep{Szegedy:2014}, and information dropping \citep{DeVries:2017, Zhong:2017}. Spatial transformation involves fundamental augmentation techniques such as random scaling, cropping, flipping, and rotation, which are widely used during training. Color distortion adjusts properties such as brightness, color balance, and saturation to mimic real-world variations \citep{Szegedy:2014}. Both spatial transformations and color distortions aim to enhance model robustness by modifying specific aspects of the training data to better reflect real-world scenarios. Information dropping techniques introduce diversity by varying visual features across training samples and are particularly effective for simulating occlusion conditions in computer vision tasks, such as visual tracking, where occlusions often result from object interactions or dynamic environments~\cite{cutout_1, cutout_2}. These techniques can also help models ignore irrelevant features and focus on constraint cues, as demonstrated by \cite{huang2020aid}, Augmentation by Information Dropping (AID) improved performance in crowded scenes by reducing the importance of appearance features.

Cutout~\citep{cutout_2} and Random Erasing~\cite{cutout_1} are two widely-used methods. Cutout masks random regions with zero pixel values using a square matrix, improving model robustness, especially when combined with other regularization techniques like dropout. Random Erasing applies random-sized masks with pixel values between 0 and 255 at random locations, conditionally applied with specified probabilities and aspect ratios to simulate partial occlusion. Region-aware Random Erasing~\cite{yang2019region} extends this idea to object detection by avoiding the deletion of important regions, such as bounding box areas.

More recent approaches, such as Hide-and-Seek~\cite{hideandseek}, partition images into grids and delete content from random cells, leaving the rest with zero pixel values. A common limitation of these methods is their random nature, which may delete informative regions and degrade performance. To address this, KeepAugment~\cite{gong2021keepaugment} uses saliency maps to identify and preserve informative areas while augmenting less important regions. Similarly, GridMask~\cite{chen2020gridmask} organizes image regions into grids and selectively masks portions, improving performance compared to AutoAugment. GridCut~\cite{feng2021gridcut} further adapts grid masks to image-specific characteristics for more nuanced augmentation, while FenceMask~\cite{li2020fencemask} applies finer, wire-like grid patterns to reduce information loss and improve generalization, especially in fine-grained recognition tasks.

In this work, we propose a new data augmentation technique, \bboxcut, which is primarily motivated by masking-based techniques. Majority of the existing masking-based approaches in the literature randomly identify regions to be masked. However, this random mask placement can significantly impact the performance of detection networks and may or may not effectively simulate the real-world masking that occurs in wheat head scenarios. In our approach, we first identify wheat heads that do not have significant overlaps with adjacent wheat heads. This is followed by the random selection of wheat heads for mask placement, with the mask color determined through histogram analysis. We demonstrate that our approach outperforms existing mask based data augmentation strategies designed to model occlusion scenarios for object detection application.

\section{Proposed Approach}

To address the problem of improving wheat head detection under occlusions, we propose a data augmentation strategy that includes a series of steps starting from removal of overlapping bounding boxes, histogram-based dominant color estimation followed by random masking for realistic occlusion simulation.

\subsection{Identification of non-overlapping boxes}
\label{sec:bbox_removal}
The first step of our approach is to identify the bounding boxes where masks should be applied. Given the nature of occlusions, as seen in Fig.~\ref{fig:OcclusionDemo}, selecting appropriate candidate bounding boxes for masking is challenging. This is crucial because applying masks to randomly selected bounding boxes may result in masking regions that are already occluded, which could severely impact detection performance. Therefore, it is essential to identify bounding boxes where the likelihood of the occlusions being present is very minimal. To achieve this, we use the intersection-over-union (IoU) metric to detect overlapping bounding boxes and exclude them from our analysis, minimizing the adverse impact on detection performance. 

Let \( B \) be the set of all the bounding boxes in an image. For each pair of bounding boxes \( b_i, b_j \in B \), we compute the intersection over Union (IoU) as:

\begin{equation}
\text{IoU}(b_i, b_j) = \frac{\text{area}(b_i \cap b_j)}{\text{area}(b_i \cup b_j)}.
\label{eq:iou}
\end{equation}

Bounding boxes with significant overlap, where \( \text{IoU}(b_i, b_j) > \tau_{\text{IoU}} \), are excluded for the masking step. The resulting subset \( B_{\text{non-overlap}} \subseteq B \) satisfies the following:

\begin{equation}
\forall b_i, b_j \in B_{\text{non-overlap}}, \quad i \neq j \implies \text{IoU}(b_i, b_j) \leq \tau_{\text{IoU}}.
\label{eq:non_overlap}
\end{equation}

\subsection{Dominant Color Estimation}
\label{sec:dominant_color_estimation}
Before the actual masking step, it is important to determine the mask color that will effectively simulate real world occlusions. Most masking-based techniques~\cite{cutout_1, cutout_2} use a fixed color mask for all training images. However, as seen in Fig.~\ref{fig:OcclusionDemo}, the color of the real-world occlusions is region-specific and depends on the environment in which the image is captured. We propose using the dominant color of the image as the mask color, as the occluders in the GWHD 2021 dataset are most likely to have colors that are dominant in the respective images.

The dominant color of the image is identified using histogram analysis, which determines the most frequently occurring intensity values across the RGB channels. Let \( I \) represent an image with \( N \) pixels. The intensity distribution of each channel \( c \in \{\text{R}, \text{G}, \text{B}\} \) is represented as a histogram \( H_c \), where \( H_c[k] \) denotes the number of pixels with intensity \( k \) in channel \( c \):

\begin{equation}
H_c[k] = \sum_{i=1}^{N} \mathbbm{1}(I_c[i] = k), \quad k \in \{0, 1, \dots, 255\},
\label{eq:histogram}
\end{equation}

where \( \mathbbm{1}(\cdot) \) is the indicator function that outputs 1 if the condition is true and 0 otherwise. The dominant intensity \( k_{\text{dom}, c} \) for each channel is defined as the intensity value with the maximum frequency:

\begin{equation}
k_{\text{dom}, c} = \arg\max_{k} H_c[k].
\label{eq:dominant_intensity}
\end{equation}

The dominant color \( C_{\text{dom}} \) is then expressed as:

\begin{equation}
C_{\text{dom}} = (k_{\text{dom}, R}, k_{\text{dom}, G}, k_{\text{dom}, B}).
\label{eq:dominant_color}
\end{equation}

This dominant color represents the objects most likely to cause occlusions in the image and is used to create the mask, enhancing the robustness of wheat head detection in real-world scenarios.

\subsection{Masking}
\label{sec:masking}

After identifying the mask color and removing the list of overlapping bounding boxes, a second probabilistic sampling is applied to \( B_{\text{non-overlap}} \), where each bounding box \( b_i \) is retained with probability \( p_m \), producing a refined subset \( B_{\text{masked}} \):

\begin{equation}
B_{\text{masked}} = \{b_i \in B_{\text{non-overlap}} : v_i \leq p_m\},
\label{eq:refined_subset}
\end{equation}

where \( v_i \sim \text{Uniform}(0, 1) \). For each bounding box \( b_i = (x_i, y_i, w_i, h_i) \in B_{\text{masked}} \), a random region \( b_{i,\text{mask}} \) within \( b_i \) is sampled for masking. The parameters \( (x_i', y_i', w_i', h_i') \) of \( b_{i,\text{mask}} \) are randomly chosen to ensure that they are fully contained within \( b_i \):

\begin{equation}
w_i' \sim \text{Uniform}(0, \alpha_w \cdot w_i), \quad h_i' \sim \text{Uniform}(0, \alpha_h \cdot h_i).
\label{eq:mask_region_2}
\end{equation}

\begin{equation}
x_i' \sim \text{Uniform}(x_i, x_i + w_i - w_i'), \quad y_i' \sim \text{Uniform}(y_i, y_i + h_i - h_i'),
\label{eq:mask_region_1}
\end{equation}

where \( \alpha_w \) and \( \alpha_h \) determine the percentage of areas to be sampled. 
The masking operation \( M \) occludes the sampled region \( b_{i,\text{mask}} \) by modifying the pixel values in that region to the dominant color \( C_{\text{dom}} \):

\begin{equation}
M(I, b_{i,\text{mask}}) = C_{\text{dom}} \cdot \mathbbm{1}_{\text{mask}(b_{i,\text{mask}})} + I \cdot (1 - \mathbbm{1}_{\text{mask}(b_{i,\text{mask}})}),
\label{eq:mask_operation}
\end{equation}

where \( \mathbbm{1}_{\text{mask}(b_{i,\text{mask}})} \) is an indicator function that assigns the dominant color to pixels inside \( b_{i,\text{mask}} \) and leaves the rest of the image unchanged. The augmented image \( I_{\text{aug}} \) is generated as:

\begin{equation}
I_{\text{aug}} = M(I, b_{i,\text{mask}}) \quad \text{for all } b_i \in B_{\text{masked}}.
\label{eq:augmented_image}
\end{equation}

\begin{figure}
	\centering
		\includegraphics[width=1\linewidth]{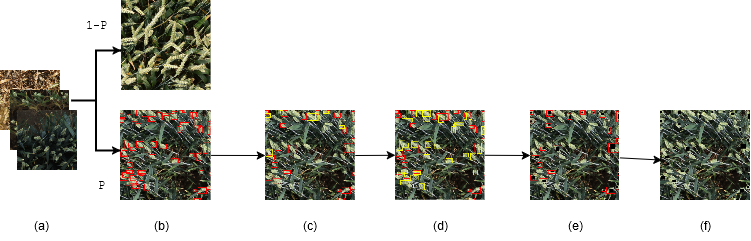}
	\caption{BboxCut Augmentation. (a) Batch of images from training set. (b) Randomly sample the images from training set to which BboxCut is to be applied. (c) Identify non-overlapping boxes using IoU metric. (d) Randomly sample onto which BboxCut needs to be applied. (e) Randomly select the region of masking based on the percentage. (f) Augmented image.    
    Color codes: Selected bounding boxes \fcolorbox{red}{red}{}, Unselected bounding boxes \fcolorbox{yellow}{yellow}{  }, BBoxcut mask \fcolorbox{black}{black}{}.}
	\label{fig:bboxcut_example}
\end{figure}

This probabilistic sampling framework, coupled with dominant color masking, systematically creates occlusion-rich training samples. The use of probabilities \( p_s \) and \( p_m \) allows fine-grained control over the augmentation process, generating diverse occlusion scenarios to improve the robustness of wheat head detection models under real-world conditions. Fig.~\ref{fig:bboxcut_example} presents a visual demonstration of our proposed approach. Note that the black mask was used for a visual demonstration. However, we use the approach described in Sec.~\ref{sec:dominant_color_estimation} to mask the randomly selected regions of the selected bounding boxes. The algorithm~\ref{alg:dominant_color_augmentation} summarizes the proposed approach using a step-by-step procedure.

\begin{algorithm}
\caption{Dominant Color-Based Masking and Augmentation}\label{alg:dominant_color_augmentation}
\begin{algorithmic}[1]
\Require Training set \( \mathcal{D} = \{I_1, I_2, \dots, I_M\} \), Bounding boxes \( B = \{b_1, b_2, \dots, b_N\} \), probabilities \(p_m \), area percentage \(\alpha_w, \alpha_h\), IoU threshold \( \tau_{\text{IoU}} \), augmentation probability \( p_{\text{aug}} \).
\Ensure Augmented dataset \( \mathcal{D}_{\text{aug}} \).

\For{each \( I_i \in \mathcal{D} \)}
    \State Sample \( r_i \sim \text{Uniform}(0, 1) \)
    \If{$r_i \leq p_{\text{aug}}$}
        \State Remove overlapping boxes from \( B \) using IoU and threshold \( \tau_{\text{IoU}} \) as in Eq.~\eqref{eq:non_overlap}.
        \State Compute histograms for each RGB channel \( H_R, H_G, H_B \) of \( I_i \) as in Eq.~\eqref{eq:histogram}.
        \State Estimate dominant color \( C_{\text{dom}} \) from histograms as in Eq.~\eqref{eq:dominant_color}.
        \State Apply probabilistic sampling to \( B_{\text{non-overlap}} \) with probability \( p_m \), resulting in \( B_{\text{masked}} \) as in Eq.~\eqref{eq:refined_subset}.
        \For{each \( b_m \in B_{\text{masked}} \)}
            \State Compute masked region \( b_{m,\text{mask}} \) as in Eqs.~\eqref{eq:mask_region_1} \& \eqref{eq:mask_region_2}.
            \State Apply masking \( M(I_i, b_{m,\text{mask}}) \) using dominant color \( C_{\text{dom}} \) as in Eq.~\eqref{eq:mask_operation}.
        \EndFor
        \State \( \mathcal{D}_{\text{aug}} \longleftarrow \mathcal{D}_{\text{aug}} \bigcup I_{i,\text{aug}} \).
        \State \textbf{continue}
    \EndIf
    \State \( \mathcal{D}_{\text{aug}} \longleftarrow \mathcal{D}_{\text{aug}} \bigcup I_{i} \).
\EndFor
\State Output \( \mathcal{D}_{\text{aug}} \).
\end{algorithmic}
\end{algorithm}

\section{Experimental Results}
\label{sec:experimental_results}

In this section, we present the quantitative and qualitative analysis of the proposed \bboxcut algorithm on the GWHD 2021 dataset. We show that our data augmentation technique can aid the wheat head localisation performance across three different kinds of object detection architectures~(\frcnn, \fcos, \detr). We perform ablation studies to understand how the parameters of \bboxcut affect the performance of the output model. We also show an improvement in the ability of the trained model to detect in the presence of occlusions.

%--------------------------------------------------------------
\subsection{Dataset}
\label{sec:dataset_description}
The Global Wheat Head Detection (GWHD) 2021 dataset~\citep{David:2021} is the largest dataset proposed for multiple phenotyping tasks. It contains 6,000 RGB images with a resolution of 1024 × 1024 pixels, collected between 2015 and 2020 by 16 institutions in 11 countries, covering genotypes from Europe, Africa, Asia, Australia, and North America. The dataset includes images taken under diverse lighting conditions and at various developmental stages, such as post-flowering and ripening, which are clustered into domains or sessions. The training set consists of 3,657 images in 18 domains. The validation set comprises data from 11 domains, totaling 1,476 images, while the test set includes data from 18 domains, totaling 1,382 images.

\subsection{Network Architecture}
\label{sec:network_arch_details}
We evaluate the performance of our data augmentation on diversified detector types namely \frcnn, \fcos and \detr. \fcos and \frcnn are single-stage and two-stage object detection architectures, respectively, while \detr is a single-stage transformer-based object detection architecture. All networks are initialized with COCO pretrained weights.  

%--------------------------------------------------------------
\subsection{Training Details}
For all detectors, we used an early stop criterion with a patience of 10 epochs. Adam (learning rate = 0.00001, batchsize=16, weight decay = 0.0001) was used as an optimizer for all experiments. ReduceLRonPlateau schedulers~\citep{ReduceLROnPlateau:2022} (factor=0.1, patience=5, threshold = 0.0001, min-lr = 0, eps = 1e-08) have been used as a learning rate scheduler. The models for \frcnn and \fcos are imported from the torchvision library, while the \detr was used from the META's official repository. All models were trained on the Colab platform. We used mean average precision (mAP) as a metric to compare performance across various detectors. From empherical analysis, we found that the best values for the constants used in our approach are as follows: \( p_{aug} = 0.3, p_m = 0.3, \alpha_w = 0.3, \alpha_h = 0.3, \tau_{\text{IoU}} = 0.5 \). 

%--------------------------------------------------------------
\subsection{Quantitative Analysis}
\label{sec:quantitative_analysis}

In this subsection, we compare the quantitative performance of the proposed approach with state-of-the-art masking-based data augmentation techniques. Although there are several masking-based approaches in the literature as described in Sect.~\ref{sec:data_augmentation}, we pick the most relevant techniques that specifically aim to model occlusions for comparison. Table~\ref{tab:Quantitative_Analysis} presents the performance of our approach in three different object detection architectures, in comparison with existing masking-based augmentation methods. The results demonstrate that our approach consistently outperforms competitive techniques, highlighting the importance of using constrained masking in complex datasets like GWHD 2021 to effectively simulate occlusion scenarios.

The most comparable method is that proposed by \cite{yang2019region}, but our approach demonstrates superior performance in all objects detection models tested. This can be attributed to the fact that \cite{yang2019region} does not consider the nature of bounding boxes, increasing the likelihood of masking already occluded wheat heads. This masking can further degrade the model’s ability to detect wheat heads correctly, leading to suboptimal performance. Furthermore, our method shows a significant improvement over the technique introduced by \cite{cutout_2}, which often applies masks randomly throughout the image, including to the background areas. This random masking does not accurately simulate the real occlusion conditions encountered in complex phenotyping scenarios like those in GWHD 2021. In contrast, our approach leverages information about the spatial distribution of wheat heads, selectively applying masks to specific regions to better mimic realistic occlusion events. In doing so, our method encourages the detection network to learn more robust features for wheat head detection, resulting in improved performance across various object detection architectures.

\begin{table}
    \centering
    \begin{tabular}{|c|c|c|c|}
    \hline
         & \frcnn & \fcos & \detr \\
    \hline
    Baseline &  51.14  & 57.18 & 57.40 \\
    \hline
    CutOut~\cite{cutout_2} & 51.29 & 57.87 & 58.20 \\
    \hline
    Region-aware Random Erasing~\cite{yang2019region} & 52.23 & 58.82 & 58.50 \\
    \hline
    \bboxcut (Ours)  & \textbf{53.90} & \textbf{60.44} & \textbf{59.30} \\
    \hline
    \end{tabular}
    \caption{Quantitative Analysis for GWHD 2021 dataset (mAP scores).}
    \label{tab:Quantitative_Analysis}
\end{table}

\subsection{Qualitative Analysis}
\label{sec:qualitative_analysis}

\begin{figure}
    \centering
    \begin{tabular}{cccc}
        \includegraphics[width=85px, height=85px]{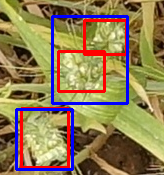}   &  
        \includegraphics[width=85px, height=85px]{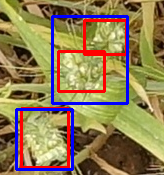} &
        \includegraphics[width=85px, height=85px]{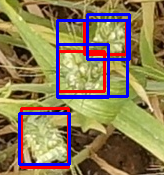} &
        \includegraphics[width=85px, height=85px]{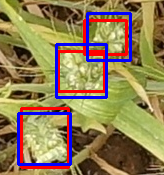} \\
        \includegraphics[width=85px, height=85px]{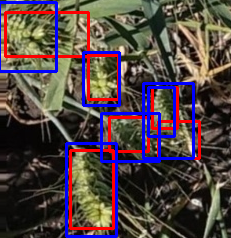}   &  
        \includegraphics[width=85px, height=85px]{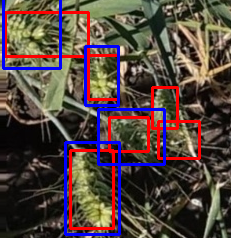} &
        \includegraphics[width=85px, height=85px]{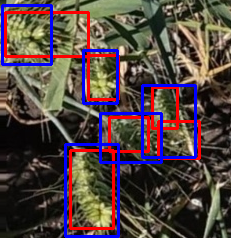} &
        \includegraphics[width=85px, height=85px]{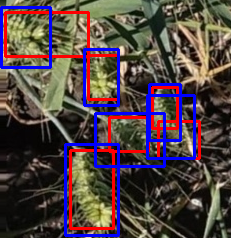} \\
        \includegraphics[width=85px, height=85px]{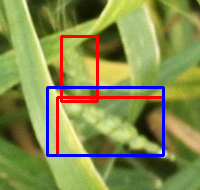}   &  
        \includegraphics[width=85px, height=85px]{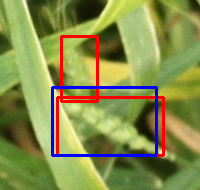} &
        \includegraphics[width=85px, height=85px]{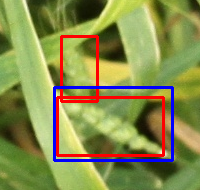} &
        \includegraphics[width=85px, height=85px]{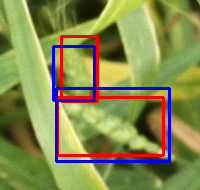} \\
        
        (a) & (b) & (c) & (d) 
    \end{tabular}
    \caption{Qualitative Analysis for \frcnn detector. Detections of: (a) Baseline, (b) Cutout, (c) Region-aware Random Erasing~\cite{yang2019region}, and (d) Proposed approach. Red box: Ground truth. Blue Box: Prediction.}
    \label{fig:qualitative_analysis}
\end{figure}

In this subsection, we present a qualitative analysis of the proposed approach compared to competitive techniques. Fig.~\ref{fig:qualitative_analysis} shows the detection results from vanilla \frcnn, Cutout, Region-Aware Random Erasing, and the proposed approach in the first, second, third, and fourth columns, respectively.  In the first row of Fig.~\ref{fig:qualitative_analysis}, the baseline and competitive techniques either mislocalize the overlapping wheat heads or produce duplicate detections, whereas the proposed approach accurately localizes them. In the second row, the proposed method successfully detects all six wheat heads, while the other techniques miss some detections. In the third row, a small wheat head partially occluded by a leaf is correctly localized using the proposed approach, whereas other methods fail to detect it. The superior performance of our method can be attributed to its controlled and adaptive masking strategy. By identifying overlapping wheat heads and strategically selecting the mask color based on the dominant image region, our approach minimizes the risk of masking already occluded wheat heads. This targeted augmentation enhances the model’s ability to learn discriminative features, resulting in more accurate localization.

\subsection{Influence on the choice of mask color}
\label{sec:mask_color_choice}

One of the key contributions of the proposed approach is the adaptive masking strategy to model occlusions. In this subsection, we analyze the impact of the chosen mask color on the performance of the detector. Some of the popular mask colors include black, white, gray, and random. However, these are generic choices and may not accurately capture the nature of occlusions in real-world scenarios. We demonstrate that an adaptive color mask performs better than existing mask color choices. Table~\ref{tab:mask_color_choice} presents the performance of the \frcnn detector with different mask color choices. The results show that the proposed approach, in the last two rows of Table~\ref{tab:mask_color_choice}, outperforms the other mask color choices, demonstrating that adaptive masking leads to improved detection accuracy by mitigating the limitations of static masks and providing a more realistic training scenario. The primary reason for this improvement is that adaptive color masks more effectively model real-world occlusions by dynamically selecting colors that blend naturally with the surrounding region. In contrast, static mask colors introduce a distribution shift between training and real-world data, as they do not accurately resemble actual occlusions such as objects partially covering one another in complex scenes. This distribution shift can lead to unnatural artifacts that hinder detection performance. In addition, generic masks can obscure valuable contextual information, making it harder for the model to learn robust features for occluded objects. By adapting to the context, the proposed approach ensures that the occlusions resemble real-world scenarios, preserving critical cues and improving generalization. Furthermore, adaptive masking enhances robustness by exposing the detector to a diverse range of occlusions, reducing its reliance on specific patterns and improving its ability to handle varying lighting conditions, textures, and object interactions. 
\begin{table}[]
    \centering
    \begin{tabular}{|c|c|}
    \hline
    Mask Color & mAP \\
     \hline
     Black   &  48.75 \\
     \hline
     Gray  & 50.51  \\
     \hline
     White   &  52.75 \\
     \hline
     Random   & 52.70  \\
     \hline
     Global Dominant Color (Ours) & \textbf{53.90} \\
     \hline
    \end{tabular}
    \caption{\frcnn performance~(mAP) on the choice of mask color.}
    \label{tab:mask_color_choice}
\end{table}

\section{Conclusions}
\label{sec:conclusion}
In this study, we proposed a constrained masking-based data augmentation technique to improve wheat head detection under occlusion conditions in the GWHD 2021 dataset. The method outperformed state-of-the-art masking-based augmentation techniques across multiple object detection architectures, both qualitatively and quantitatively, indicating that our approach is agnostic to the detection network. This also demonstrates that our method effectively simulates realistic occlusion scenarios by carefully identifying wheat heads with minimal overlap and selectively applying masks based on histogram analysis, ensuring meaningful data transformations. In general, this work highlights the importance of incorporating domain-specific constraints into data augmentation strategies to improve object detection performance. In future research, we plan to extend our approach to other datasets and explore adaptive, multistage augmentation techniques for further improvements in occlusion modeling and domain generalization.

\section*{Acknowledgments}
This work has been partially supported by Research England (Lincoln Agri-Robotics) as part of the Expanding Excellence in England (E3) Program and the School of Computer Science, University of Lincoln, UK. The views expressed in this publication are solely those of the authors and not necessarily those of the funders.

\section*{Declaration of generative AI and AI-assisted technologies in the writing process}
During the preparation of this work, the author(s) used ChatGPT and Writefull to improve the readability and language of the work. After using this tool/service, the author(s) reviewed and edited the content as needed and takes (s) full responsibility for the content of the publication.



\bibliographystyle{elsarticle-num} 
\bibliography{sample}


%\begin{thebibliography}{00}


%\bibitem{}

%\end{thebibliography}
\end{document}